\documentclass{article}
\pdfoutput=1
\usepackage{ijcai21}

\usepackage{times}
\usepackage{soul}
\usepackage{url}
\usepackage[hidelinks]{hyperref}
\hypersetup{colorlinks,allcolors=black}
\usepackage[utf8]{inputenc}
\usepackage[small]{caption}
\usepackage{graphicx}
\usepackage{amsmath}
\usepackage{amsthm}
\usepackage{booktabs}
\usepackage{algorithm}
\usepackage{algorithmic}
\title{Reducing Discrimination in Learning Algorithms for Social Good in Sociotechnical Systems}

\author{
    Katelyn Morrison
    \affiliations
    Department of Computer Science, University of Pittsburgh, United States
    \emails
    katelyn.morrison@pitt.edu
}

\begin{document}

\maketitle

\begin{abstract}
Sociotechnical systems within cities are now equipped with machine learning algorithms in hopes to increase efficiency and functionality by modeling and predicting trends. Machine learning algorithms have been applied in these domains to address challenges such as balancing the distribution of bikes throughout a city and identifying demand hotspots for ride sharing drivers. However, these algorithms applied to challenges in sociotechnical systems have exacerbated social inequalities due to previous bias in data sets or the lack of data from marginalized communities. In this paper, I will address how smart mobility initiatives in cities use machine learning algorithms to address challenges. I will also address how these algorithms unintentionally discriminate against features such as socioeconomic status to motivate the importance of algorithmic fairness. Using the bike sharing program in Pittsburgh, PA, I will present a position on how discrimination can be eliminated from the pipeline using Bayesian Optimization.
\end{abstract}

\section{Introduction}
As smart cities emerge, sociotechnical systems are starting to eliminate traditional systems such as replacing transportation systems with advanced technologies \cite{gatech} - also known as smart mobility systems. These initiatives are spreading from city to city, introducing dynamic, on-demand bike sharing programs, ride hailing programs, and car sharing programs \cite{yan2020fairness}. Emerging smart mobility systems are supported by a variety of Information and Communication Technologies \cite{benevolo2016smart} which have allowed for the possibility of collecting copious amounts of data on citizens' mobility patterns. Citizens can now interact with these systems through hardware and software leaving behind traces of their data and insights about their behaviors in the future \cite{rangwalacheck}. 

 People interact with a variety of smart devices connected to the internet that provide services such as navigation kiosks or apps, bus routing apps, smart card readers, and ride sharing apps numerous times throughout their day to navigate the city generating a vast amount of valuable data. This data makes up a large data set that is then used in learning algorithms to assist urban planners, policy makers, and government officials in evaluating policies or making decisions. In terms of smart mobility initiatives, the data helps decision makers evaluate where to redistribute resources and when to eliminate resources in an area with little demand \cite{yan2020fairness,luo2020d3p}. Nonetheless, these complex, dynamic systems collecting copious amounts of data, and the citizens providing the data, are increasingly important agents in the decision making and policy making processes. Thus, citizens play an important role within several systems in the city including the transportation system and the future governance related to it.

Smart mobility initiatives are motivated by providing a cheap, yet dynamic, on-demand mode of transportation for citizens \cite{yan2020fairness}. Thus, the accessibility and availability of smart mobility options are vital within a city. Furthermore, the fairness of the algorithms used for these initiatives are equally important to avoid intensifying inequalities already present in a given area. Initiatives such as bike sharing programs have the potential to increase the overall quality of life and health of citizens \cite{mooney2019freedom}. Other initiatives such as car sharing or ride hailing have the potential to reduce traffic congestion which ultimately helps reduce the CO$_2$ emissions \cite{luo2020d3p}. Furthermore, these affordable, dynamic, on-demand modes of transportation help connect citizens to the job market and economy within cities. These smart mobility initiatives can only fulfill their mission if they are equally accessible to all social classes and all neighborhoods within a city. 

Learning algorithms have been used to complement smart mobility initiatives, but the algorithms focus on past usage data to predict future trends. However, it has been observed that using past usage data tends to intensify any social inequalities that are present in the previous usage data \cite{chui_harrysson_manyika_roberts_chung_nel_heteren_2019}. Repercussions of the current techniques will be revealed through several related works that focus on smart mobility systems and fairness in machine learning algorithms. 

\section{Related Work}

The awareness of algorithmic bias and lack of fairness in learning algorithms is not novel. Recently, the notion of fairness in learning algorithms applied to urban systems has been explored. However, research has highlighted the lack of previously acknowledging the impact these learning algorithms have had on mobility systems \cite{yan2020fairness}. There are several works considering novel techniques to predict demand of smart mobility systems or identify hotspots at a given space and time, but as \cite{yan2020fairness} mentioned, "No existing work in modeling urban resource demand considers fairness in their solutions". The following works that consider or show a lack of considering fairness within their algorithm provide a strong motivation that learning algorithms need to be reevaluated in the context of urban and social systems. 

\subsection{Applications in Sociotechnical Systems} 

\cite{yan2020fairness} develops a model that represents the demand prediction of bikes in a bike sharing program or demand hotspots of rides for a ride hailing driver. Specifically, this model is based on three convolutional neural networks (CNN) combined into one to model the demand of smart mobility resources. They use regularizers in their loss function to ensure fairness while training and testing the model. In their work, they define fairness, "... as the requirement that individuals of different demographic groups receives equal amount of mobility resources".

Another study looks at the expansion of electric vehicle sharing systems and uses a data-driven approach to predict the demand of the system currently, in the near future, and in the long term \cite{luo2020d3p}. They use features from a station while considering the global spatial and local temporal attributes to a given system's station network. Their demand prediction model is made up of Graph Convolutional Neural Networks (GCN), but does not strongly consider fairness of the prediction given that they are incorporating historical knowledge.

Two recent studies evaluate fairness, but in a slightly different context. Switching from the customer's perspective to the provider's, or driver's, perspective, these two studies focus on the fairness of the drivers' income or opportunity to pick up passengers \cite{rong2019tesla,lesmana2019balancing}. \cite{rong2019tesla} specifically studies this problem in the context of Changsha, China to optimize the dispatching of taxis to predicted hotspots fairly. \cite{lesmana2019balancing} focuses more broadly on the ride hailing system instead of specifically taxis and evaluate their contributions on data from New York City. They both want to optimize the system's efficiency while incorporating the fairness of revenue among drivers and acknowledge its potential to be applied in other domains relating to sociotechnical systems.  

\subsection{Fairness in Learning Algorithms}
Learning algorithms used in the domain of smart mobility have provided significant assistance to decision makers and urban planners, but they do not incorporate fairness. Creating a classifier that incorporates fairness effectively will not be achieved by simply abstracting out features or ignoring sensitive attributes \cite{barocas-hardt-narayanan}. \cite{barocas-hardt-narayanan} presents a list of several proposed fairness criteria - the below works propose classifiers that satisfy statistical parity.

\cite{zemel2013learning} creates a fair classifier by generating a probabilistic mapping where necessary information about individuals is mapped from the original distribution to another distribution. Furthermore, information of a given protected feature is dissociated from that individual in the new distribution. They achieve this by having their classifier learn a set that satisfies group fairness (i.e. statistical parity or demographic parity).

Another approach creates a classifier by adjusting how similar individuals should be treated in the given classification task \cite{dwork2011fairness}. Similar to \cite{zemel2013learning}, they investigate how their classifier can satisfy statistical parity. Their classifier satisfies statistical parity under the \textit{Lipschitz} condition. 

\subsection{Evaluating Accessibility in Systems}
Some studies have evaluated and investigated the accessibility, fairness, and discrimination present within current smart mobility systems. Two studies in Manizales, Colombia investigate the social accessibility of the public bike sharing system, \textit{Manizales En Bici}. An important characteristic of Manizales is that it is split into sectors, or neighborhoods, that are grouped by socioeconomic status. One study investigates the problem of unequal distribution of the bikes by calculating the minimum travel time to get around the city in comparison to socioeconomic status and geography \cite{cardona2017analisis}. Another study uses ArcGIS software \cite{zuluagapropuesta} to visualize this. Overall, both of these studies agree that the locations of the \textit{Manizales En Bici} stations are more accessible or available to individuals of higher socioeconomic status.

A separate study evaluated another bike sharing program in Seattle, WA, where the bikes are not required to be returned to a station \cite{mooney2019freedom}. This study discovers that while no neighborhood was being exclusively avoided, the richer neighborhoods tended to have more bikes available. Their conclusion from the studies pointed out that dock-less, or station-less, smart mobility systems can potentially offer more equity than other models.

\section{Preliminary Work \& Methods}

I analyzed the Pittsburgh's Healthy Ride bike sharing program to identify the fairness, or accessibility, of the current distribution of bike stations. Accessibility can be determined by the relationship between the number of bike stations, the moving average number of bikes at a given station, and related demographic and economic features in a given census tract. The geospatial analysis in \textbf{figure 1} reveals that 95\% of the bike stations were placed in neighborhoods that were considered to have little to no evidence of poor housing conditions. 
\begin{figure}[h]
\centering
\includegraphics[width=0.9\linewidth,height=5cm]{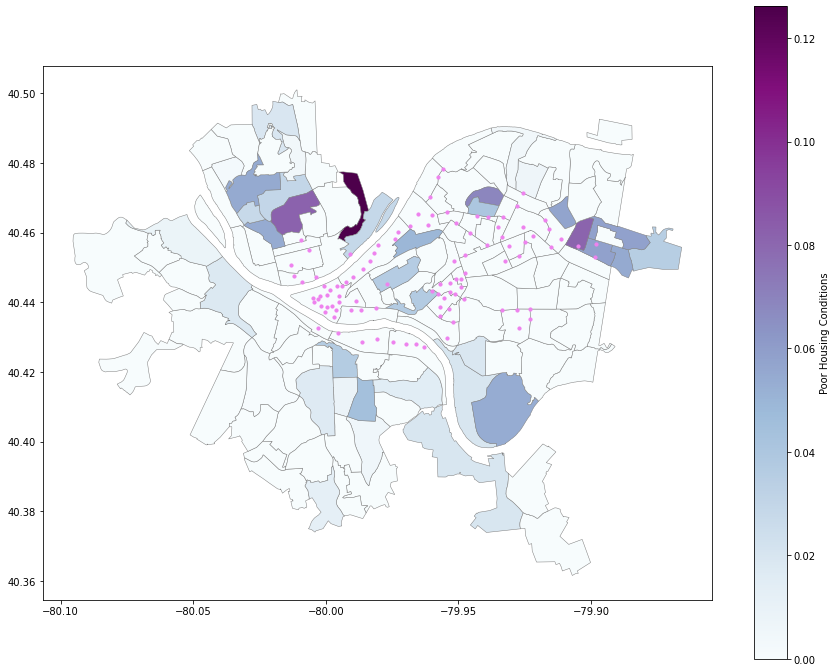}
\caption{Mapping bike station locations to poor housing conditions in Pittsburgh, PA, USA. Data from \protect\cite{jones,khalil}}
\end{figure} 
The geospatial analysis in \textbf{figure 2} reveals the median household income in 2015 in relation to the bike stations. One cluster of bike stations circled in magenta identifies outliers in the income data: this region of Pittsburgh is primarily student housing where students rent or sublease houses. This area also broadly encompasses three universities in Pittsburgh. These two figures help set the stage for the Bayesian Optimization approach that I propose.

In the context of smart mobility systems, it is common to use supervised learning algorithms to predict demand, identify demand hotspots, and identify the best location to distribute resources to help decision makers and urban planners. These algorithms are applied to help mitigate the exploitation vs. exploration problem that occurs in bike sharing programs. Considering the Pittsburgh Healthy Ride bike sharing program, I propose a novel approach to identify demand hotspots and prime locations while trying to incorporate fairness similar to \cite{zemel2013learning} by dissociating information of selected features from a neighborhood or census tract. 

Pittsburgh's bike sharing program has 100 bike stations which have been placed based on 6 criteria defined by the Healthy Ride organization. They define a good location for a station to be within $1/4$ mile from another station, on public land, not obstructing other public utilities, not hidden in allies or side streets, has a certain amount of square footage to build the station, and has access to sunlight \cite{healthyridepgh}. Their current plan for expansion is undetermined and they are asking for users to fill out a feedback form. Soliciting feedback for this system can result in profound disparate impact \cite{barocas-hardt-narayanan}. With this is mind, I propose modifying this current decision making process by using Bayesian Optimization for model-based reinforcement learning to explore new areas for stations followed by Bayesian inference to update the prior and posterior based on evidence that results from the exploration. To satisfy statistical parity, the data can be mapped to a new distribution \cite{zemel2013learning}.

In the case of bike sharing programs, the agent is a planner and they are trying to optimize the location to put a bike station or the location to reallocate current bikes to another station. It is costly and takes time to build the bike stations, thus the planners want to make sure that they will build the station in a location that will have a high demand. The planner is faced with an exploration vs. exploitation trade off. An agent (i.e. planner) would be rewarded every time it explores uncharted neighborhoods or neighborhoods with very little historical user data. These rewards are to encourage the planner to explore and identify regions that do not have any source of historical data to contribute to evidence of demand. The policies for the agent (i.e the planner) will represent the probabilities for each action that can be taken (i.e. the next neighborhoods a planner should construct a station at). These probabilities would be calculated and updated by using an acquisition function.

\begin{figure}[h]
\centering
\includegraphics[width=0.9\linewidth,height=5cm]{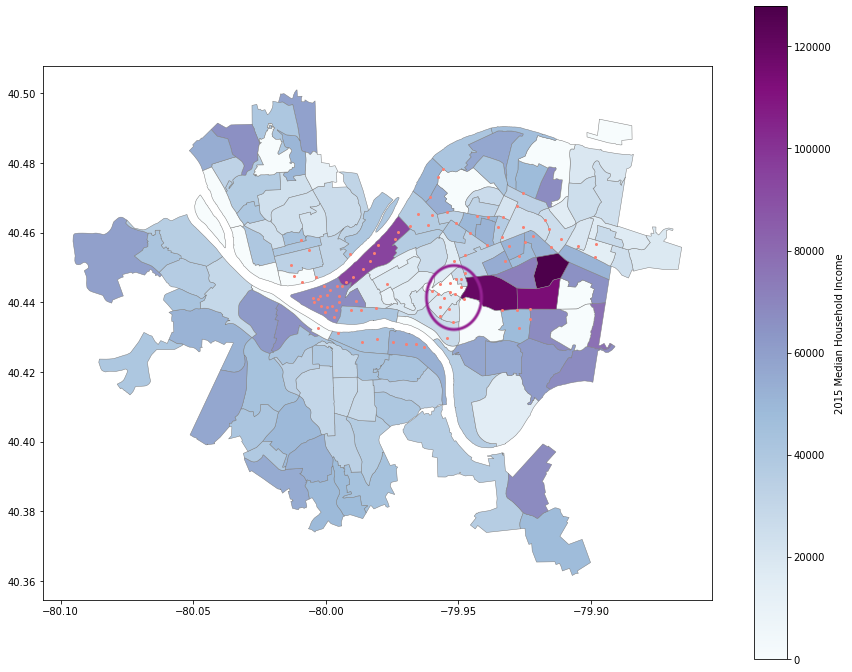}
\caption{Bike station locations and 2015 median household income in Pittsburgh, PA, USA. Data from \protect\cite{ffiec,khalil}}
\end{figure} 

The value function can be defined as the posterior using past usage data (i.e. the evidence), the prior belief (i.e previous likelihood of demand), and the current likelihood given the location the planner chooses. Currently, bike sharing programs tend to place bikes and stations in regions where there is a high certainty that demand will be high. If they ignore this high certainty, they will eventually place bikes or stations in regions that could have otherwise been excluded or discriminated against. This approach allows planners to consider evidence of past user demand for certain locations and times. It also allows for the demand of station locations in previously unexplored regions to be identified. This approach forces the bike sharing programs to explore new regions and potentially identify a new cohort of users for their system. 

To execute this, several features need to be integrated into the algorithm so that an unexplored area in the city that has no points of interest is not chosen with a high confidence. For example, if an entire census tract consisted of forested land, no housing, no businesses, and no points of interest, it would not be beneficial to have the algorithm suggest this area. Thus, spatial features will be needed such as point of interests (i.e. businesses, housing developments, and other transit stations). 

\section{Conclusion \& Discussion}
When considering future work in fairness-aware learning algorithms applied in sociotechnical systems, researchers should avoid developing and deploying algorithms that adhere to "fairness through unawareness" \cite{DBLP:journals/corr/HardtPS16}. Identifying alternatives to the current demand and hot spot prediction algorithms are imperative to increasing the accessibility of smart mobility options to all citizens. There are numerous contributions progressing fairness of machine learning algorithms. There are also numerous contributions integrating learning algorithms in sociotechnical systems to achieve efficiency, but there is a lack of contributions progressing fairness of learning algorithms in urban systems. Future work should consider novel approaches that combine novel techniques from both domains to reduce discrimination. I identify an approach using Bayesian optimization for model-based reinforcement learning to help planners identify locations to offer smart mobility systems.

\newpage

\bibliographystyle{named}
\bibliography{ijcai21}

\begin{thebibliography}{}

\bibitem[\protect\citeauthoryear{Barocas \bgroup \em et al.\egroup
  }{2019}]{barocas-hardt-narayanan}
Solon Barocas, Moritz Hardt, and Arvind Narayanan.
\newblock {\em Fairness and Machine Learning}.
\newblock fairmlbook.org, 2019.
\newblock \url{http://www.fairmlbook.org}.

\bibitem[\protect\citeauthoryear{Benevolo \bgroup \em et al.\egroup
  }{2016}]{benevolo2016smart}
Clara Benevolo, Renata~Paola Dameri, and Beatrice D’auria.
\newblock Smart mobility in smart city.
\newblock In {\em Empowering Organizations}, pages 13--28. Springer, 2016.

\bibitem[\protect\citeauthoryear{Cardona \bgroup \em et al.\egroup
  }{2017}]{cardona2017analisis}
Mateo Cardona, Juan Zuluga, and Diego Escobar.
\newblock An{\'a}lisis de la red de ciclo-rutas de manizales (colombia) a
  partir de criterios de accesibilidad territorial urbana y cobertura de
  estratos socioecon{\'o}micos.
\newblock {\em Revista Espacios}, 38(28), 2017.

\bibitem[\protect\citeauthoryear{Chui \bgroup \em et al.\egroup
  }{2019}]{chui_harrysson_manyika_roberts_chung_nel_heteren_2019}
Michael Chui, Martin Harrysson, James Manyika, Roger Roberts, Rita Chung,
  Pieter Nel, and Ashley~van Heteren.
\newblock Applying artificial intelligence for social good, Nov 2019.

\bibitem[\protect\citeauthoryear{Dwork \bgroup \em et al.\egroup
  }{2011}]{dwork2011fairness}
Cynthia Dwork, Moritz Hardt, Toniann Pitassi, Omer Reingold, and Rich Zemel.
\newblock Fairness through awareness, 2011.

\bibitem[\protect\citeauthoryear{FFIEC}{2020}]{ffiec}
FFIEC.
\newblock 2020 ffiec census report - summary census income information, 2020.

\bibitem[\protect\citeauthoryear{Garcia \bgroup \em et al.\egroup
  }{2017}]{zuluagapropuesta}
Zuluaga Garcia, Juan David, et~al.
\newblock {\em Propuesta metodologica para el diagnostico y planificacion
  urbana de una red de ciclorrutas. Caso estudio: Manizales}.
\newblock PhD thesis, Universidad Nacional de Colombia-Sede Manizales, 2017.

\bibitem[\protect\citeauthoryear{Hardt \bgroup \em et al.\egroup
  }{2016}]{DBLP:journals/corr/HardtPS16}
Moritz Hardt, Eric Price, and Nathan Srebro.
\newblock Equality of opportunity in supervised learning.
\newblock {\em CoRR}, abs/1610.02413, 2016.

\bibitem[\protect\citeauthoryear{HealthyRide}{2020}]{healthyridepgh}
HealthyRide.
\newblock Station planning, 2020.

\bibitem[\protect\citeauthoryear{Jones}{2020}]{jones}
Lynda Jones.
\newblock Allegheny county poor housing conditions, 2020.

\bibitem[\protect\citeauthoryear{Khalil}{2020}]{khalil}
Sara Khalil.
\newblock Healthy ride stations, 2020.

\bibitem[\protect\citeauthoryear{Lesmana \bgroup \em et al.\egroup
  }{2019}]{lesmana2019balancing}
Nixie~S Lesmana, Xuan Zhang, and Xiaohui Bei.
\newblock Balancing efficiency and fairness in on-demand ridesourcing.
\newblock In {\em Advances in Neural Information Processing Systems}, pages
  5309--5319, 2019.

\bibitem[\protect\citeauthoryear{Levine}{2020}]{gatech}
Ben Levine.
\newblock Data-driven research aims to solve first/last mile problem.
\newblock
  \url{https://www.govtech.com/transportation/Data-Driven-Research-Aims-to-Solve-First-Last-Mile-Problem.html},
  March 2020.

\bibitem[\protect\citeauthoryear{Luo \bgroup \em et al.\egroup
  }{2020}]{luo2020d3p}
Man Luo, Bowen Du, Konstantin Klemmer, Hongming Zhu, Hakan Ferhatosmanoglu, and
  Hongkai Wen.
\newblock D3p: Data-driven demand prediction for fast expanding electric
  vehicle sharing systems.
\newblock {\em Proceedings of the ACM on Interactive, Mobile, Wearable and
  Ubiquitous Technologies}, 4(1):1--21, 2020.

\bibitem[\protect\citeauthoryear{Mooney \bgroup \em et al.\egroup
  }{2019}]{mooney2019freedom}
Stephen~J Mooney, Kate Hosford, Bill Howe, An~Yan, Meghan Winters, Alon Bassok,
  and Jana~A Hirsch.
\newblock Freedom from the station: Spatial equity in access to dockless bike
  share.
\newblock {\em Journal of transport geography}, 74:91--96, 2019.

\bibitem[\protect\citeauthoryear{Rangwala}{2018}]{rangwalacheck}
Harsh Rangwala.
\newblock Socio-technical systems, Dec 2018.

\bibitem[\protect\citeauthoryear{Rong \bgroup \em et al.\egroup
  }{2019}]{rong2019tesla}
Huigui Rong, Qun Zhang, Xun Zhou, Hongbo Jiang, Da~Cao, and Keqin Li.
\newblock Tesla: A centralized taxi dispatching approach to optimizing revenue
  efficiency with global fairness.
\newblock {\em In KDD Workshop on Urban Computing (UrbComp ’20)}, 2019.

\bibitem[\protect\citeauthoryear{Yan and Howe}{2020}]{yan2020fairness}
An~Yan and Bill Howe.
\newblock Fairness-aware demand prediction for new mobility.
\newblock In {\em Proceedings of the AAAI Conference on Artificial
  Intelligence}, volume~34, pages 1079--1087, 2020.

\bibitem[\protect\citeauthoryear{Zemel \bgroup \em et al.\egroup
  }{2013}]{zemel2013learning}
Rich Zemel, Yu~Wu, Kevin Swersky, Toni Pitassi, and Cynthia Dwork.
\newblock Learning fair representations.
\newblock In {\em International Conference on Machine Learning}, pages
  325--333, 2013.

\end{thebibliography}

\end{document}